\documentclass[a4paper,twocolumn]{article}
\usepackage{jsaiac}
\usepackage{color}
\usepackage{comment}
\usepackage{url}
\usepackage[dvipdfmx]{xcolor}
\usepackage[dvips]{graphicx}

\title{\LARGE \bf Deep Predictive Learning: Motion Learning Concept\\
\LARGE \bf inspired by Cognitive Robotics}

\address{Tetsuya Ogata, Waseda University, 3-4-1 Ookubo, Shinjuku-ku, Tokyo, 169-8555, Japan, {\tt\small ogata@waseda.jp}\\
\hspace{-2.5mm}${\dagger}$ All authors have contributed equally to this work.
}

\author{%
Kanata Suzuki$^{\dagger}$\first\second
\and
\hspace{6mm}
\and
Hiroshi Ito$^{\dagger}$\first\third
\and
\hspace{6mm}
\and
Tatsuro Yamada$^{\dagger}$\first\fourth
\and
\hspace{6mm}
\and
Kei Kase$^{\dagger}$\first
\and
\hspace{6mm}
\and
Tetsuya Ogata$^{\dagger}$\first
}

\affiliate{
\first{} Waseda University
\and
\second{} Fujitsu Limited
\and
\third{} Hitachi, Ltd.
\and
\fourth{} Panasonic Corporation
}

\begin{abstract}
Bridging the gap between motion models and reality is crucial by using limited data to deploy robots in the real world. Deep learning is expected to be generalized to diverse situations while reducing feature design costs through end-to-end learning for environmental recognition and motion generation. However, data collection for model training is costly, and time and human resources are essential for robot trial-and-error with physical contact. We propose “Deep Predictive Learning,” a motion learning concept that predicts the robot’s sensorimotor dynamics, assuming imperfections in the prediction model. The predictive coding theory inspires this concept to solve the above problems. It is based on the fundamental strategy of predicting the near-future sensorimotor states of robots and online minimization of the prediction error between the real world and the model. Based on the acquired sensor information, the robot can adjust its behavior in real time, thereby tolerating the difference between the learning experience and reality. Additionally, the robot was expected to perform a wide range of tasks by combining the motion dynamics embedded in the model. This paper describes the proposed concept, its implementation, and examples of its applications in real robots.
The code and documents are available at: \url{https://ogata-lab.github.io/eipl}
\end{abstract}

\def\BibTeX{{\rm B\kern-.05em{\sc i\kern-.025em b}\kern-.08em%
 T\kern-.1667em\lower.7ex\hbox{E}\kern-.125emX}}
\def\JBibTeX{\leavevmode\lower .6ex\hbox{J}\kern-0.15em\BibTeX}
\def\LaTeXe{\LaTeX\kern.15em2$_{\textstyle\varepsilon}$}

\begin{document}
\maketitle

\section{Introduction}

"The Moravec Paradox" is currently one of the significant challenges in artificial intelligence technology \cite{Moravec1988mind}. This paradox refers to the contradiction in which tasks that even children can perform semi-unconsciously are exceedingly difficult for the latest artificial intelligence and robots compared to complex logical reasoning tasks, such as puzzles or chess. Tasks involving independent movements such as walking or running made significant technological advancements in the 2000s \cite{Garcia1996,Hirai1998}. In contrast, perception tasks such as vision and voice recognition saw substantial progress in the 2010s through deep learning techniques~\cite{Krizhevsky2012nips,He2016cvpr,Ren2016pami,Redmon2016cvpr,Hinton2013spm,Graves2014icml}. However, performing a variety of tasks using a single common hand, just like humans, remains extremely challenging even with the latest technology available today.

\subsection{Experience-based Robotics}

Deep learning methods are expanding in the environmental recognition and motion generation fields. Examples include the extraction of graspable regions of objects~\cite{Pinto2016icra,Mahler2017dex,Mousavian2019iccv} and the combinatorial exploration of action patterns through reinforcement learning~\cite{Levine2016jmlr,Levine2017ijrr}. These research examples represent an inductive approach for directly acquiring models of the environment and trajectory generation from data. In robot learning, data can be generated or acquired through bottom-up exploratory actions. Sensory-motor information obtained from active interaction, or "experience," plays an important role in robot learning. For example, while conventional learning data are annotated with labels provided by humans, experience can be utilized in a robot's motor-skill learning, even in the absence of teacher labels. Another important perspective from this concept, called experience-based robotics, is the functional separation between environmental perception and motion generation. Considering the common end-to-end approach for processing images, audio, and natural language using deep learning, it is evident that a clear separation of these two functions could also be a subject of learning.

The concept of recognizing and generating partial "functions" of autonomous systems here is important, as it is believed to emerge gradually from the interaction of the body, environment, neural circuits, and the continuous dynamics of end-to-end learning, as proposed in cognitive developmental robotics~\cite{Asada2015} or brain-inspired learning~\cite{Haruno2001}. In particular, to deploy robots in the real world, it is crucial to bridge the gap between motion models and reality using limited data. Although deep learning-based approaches show improvements in task accuracy through large amounts of data and trials, data collection for model training is costly. In particular, robot motion data depend on the hardware, making accessing extensive data for model tuning uncommon. Although research has been conducted to transfer learning results from simulation environments to the real world, modeling nonlinear physical phenomena remains challenging and unsolvable. It is necessary to achieve a wide range of task behaviors by creatively combining representations derived from the limited motion experience of robots.

\subsection{Proposed Robot Learning Concept}

This study discusses the concept of "deep predictive learning," a methodology for applying deep learning techniques to robotics, along with its implementation methods. This concept, inspired by predictive coding theory~\cite{Rao1999, Friston2006jpp, Friston2010nrn}, is based on the fundamental strategy of predicting the near-future sensory-motor states of robots and online minimization of the prediction error between the real world and the model. Furthermore, by embedding the temporal relationships between sensory and motor information in systems where the environment and body interact dynamically, the system transitions to other motion dynamics based on prediction errors during generation, thereby enabling the execution of high-dimensional, long-term tasks.

Section 2 discusses previous research and the challenges related to applying deep learning in robotics and describes the positioning of deep predictive learning. Section 3 covers the background of deep predictive learning, including predictive coding and the free-energy principle, and outlines strategies for applying them to robot motion learning. We also discuss specific implementation approaches based on deep learning techniques and examine them from the perspective of cognitive robotics, including dynamic cognitive and action-normative systems. Section 4 summarizes the published studies we developed using the proposed concept from two perspectives: scalability by combining multiple motions and application to complex manipulation tasks. Finally, in Section 5, we summarize and discuss prospects.

\section{Deep Learning in Robotics Fields}

In this section, we review previous research on the application of deep learning in robotics and establish the context for our proposed deep predictive learning concept.

\subsection{Recognition}

The term "deep learning" was introduced in robotics at the IEEE/RSJ IROS 2017 conference, which was relatively late compared with other domains. Several early applications of deep learning in robotics focused on object detection in robot vision~\cite{Ren2016pami,Redmon2016cvpr} and conversational speech recognition~\cite{Hinton2013spm,Graves2014icml}. In traditional robotic intelligence design, the process typically involves the following steps: (1) recognizing the environment or objects, (2) correlating the recognition results with pre-provided 2D maps or 3D CAD models, (3) understanding the current positions, object positions, and orientations, and (4) planning actions based on the situation. Deep learning significantly enhances the recognition capability in the first step. Learning labels, such as graspable regions that were previously challenging to model, made recognizing a wide variety of general objects possible.

However, the burden of manual labeling by humans is substantial, and relying solely on image-based learning does not guarantee successful object grasping. The grasping performance depends on a combination of visual information, object attributes, and physical characteristics of the robot. Factors such as the gripper force, friction conditions, and object deformations must be considered, as well as countless elements.

\subsection{Robot Motion Learning}

In response to the developments mentioned above in perception systems, active research has been conducted ~\cite{Noda2014multimodal,Lee2019icra,lin2022tactilegym2} on end-to-end methods for predicting robot motion trajectories directly from multimodal input sensor data (such as visual, auditory, tactile, etc.). Deep learning-based methods can be broadly categorized into two groups within learning-based approaches to robot motion.

\subsubsection{Reinforcement Learning}

In deep reinforcement learning, the agent predicts rewards for combinations of high-dimensional inputs related to the environment and several executable action outputs from the robot. With each attempt by the agent, the environment provides rewards, and the model updates its parameters to acquire a policy that maximizes the rewards~\cite{Levine2016jmlr,Levine2017ijrr,Haarnoja2017icml}. The application of deep reinforcement learning in games has been extremely successful~\cite{Mnih2013playing,Hessel2018rainbow}, and its application in robotics is a natural progression. Its greatest appeal lies in significantly reducing the traditional process of designing models for the environment or objects that designers must provide. This innovation has led to various task-specific methods~\cite{Tsurumine2019ras,Mahler2019learning}. Levine et al. replaced the previously required image feature engineering with deep learning and built an end-to-end model from images to actions~\cite{Levine2017ijrr}. However, in~\cite{lee2023pi,rlscale2023arxiv}, efforts were made to enhance the performance of general agents by training multiple tasks with a single model. However, achieving an optimal model for online learning in a real environment remains a challenge, as it requires many trials and resources. There is also an approach for collecting a large amount of data using human resources~\cite{dasari2019robonet,rt12022arxiv}, but this approach has limitations regarding data diversity.

The Sim2Real approach uses simulations to address these issues~\cite{James2019cvpr,Chebotar2019icra}. Making simulations closer to reality~\cite{allevato2020tunenet} and transforming them~\cite{rao2020rl} can supplement the lack of data. Domain randomization, especially regarding visual and physical parameters~\cite{Tobin2017}, is utilized in tasks such as multi-fingered hand manipulation~\cite{openai2018learning} and rough terrain walking of quadruped robots~\cite{lee2020learning}. In~\cite{lee2020learning}, teacher-student learning using environmental information obtained only in simulations was successful in real-world motion control, which is quite intriguing. Recent advancements in physical simulation technology can replicate scenes equivalent to the real world; however, detailed modeling of contact states remains a challenge~\cite{drake,Narang2022rss}. This problem is currently being discussed as a subject in Surface Science.

Moreover, traditional reinforcement learning typically requires pre-designing reward functions, which poses challenges because the model's performance depends heavily on it. Inverse reinforcement learning introduces a process for estimating the reward functions from human demonstrations~\cite{Peng2018,Tsurumine2019humanoids,Wang2021}, thereby reducing the burden of manually designing rewards based on task domain knowledge. However, aspects such as policy behavior design remain, and the flexibility of the tasks applied to real machines is limited.

\subsubsection{Behavioral Cloning}

Recently, there has been growing interest in offline learning to address the challenges of data scale arising from online policy updates in conventional reinforcement learning. In Behavioral Cloning (or Imitation Learning), learning is conducted to replicate the motion sequence data (joint angles, velocities, torques, etc.) obtained from robots~\cite{yu2018one,Zhao2023rss,jang2021bc,Lynch2022,guhur2023instruction,chi2023diffusionpolicy}. Zhao et al. used a low-cost fine-grained teleoperation system and employed transformers to enable learning across various tasks~\cite{Zhao2023rss,fu2024mobile}. In this framework, it is possible to acquire a motion generation model directly from the motion experience if effective data for task execution are available. In addition, the feature of not requiring a complex reward design is highlighted. Consequently, it has been applied to various high-difficulty tasks, such as flexible object manipulation~\cite{Kawaharazuka2019icra} and human-cooperative tasks~\cite{Sasagawa2020ral}.

Regarding related technologies, we should also mention offline reinforcement learning. Offline reinforcement learning aims to acquire a good policy solely from a pre-collected dataset, and these studies can be interpreted as reward-conditioned Behavioral Cloning~\cite{actionablemodels2021arxiv,ma2022vip}. These approaches seek to improve the learning efficiency by utilizing experience data~\cite{kalashnikov2018qt} or by simultaneously collecting data from multiple robots~\cite{kalashnikov2021mt}. In particular, Chen et al. proposed a sequentially predictable offline reinforcement learning model using transformers~\cite{chen2021decision}. They predicted actions from a sequence of tokens representing rewards, states, and actions and achieved high performance in long-term tasks using expected cumulative future rewards.

Many of the studies mentioned thus far have focused solely on modeling robot action sequences. In contrast, the "Predictive Learning" approach aims to perform self-supervised learning, including the external environment~\cite{lotter2016deep}. Lotter et al. proposed PredNet, which mimics the predictive coding processes in the cerebral cortex. PredNet learns to minimize the error between the predicted and actual values in video prediction tasks. These are implementations of predictive coding theory~\cite{Rao1999} using deep learning models, but they do not consider interaction with the external environment in robotics.

Our proposed concept, from the perspective of cognitive robotics~\cite{Asada2015}, incorporates predictive coding into sensorimotor dynamics. The crucial idea in our proposal is that the learning target of autonomous systems should not be a "sensor-to-motion map" but rather a "continuous loop of integrated sensorimotor dynamics." The distinction between sensation and motion is always ambiguous in systems where the environment and body interact dynamically. A continuous flow exists in which the current leads to future outcomes. It is crucial to have an orientation that fills the error of the predictive flow, which never completely fills the self and the external. This type of predictive learning system (forward model) and the corresponding motion generation system (inverse model) have been emphasized for a long time. However, practical robots designed to work with this concept in the real world remain limited.

\begin{figure}[t]
\centering
\includegraphics[width=1.0\columnwidth]{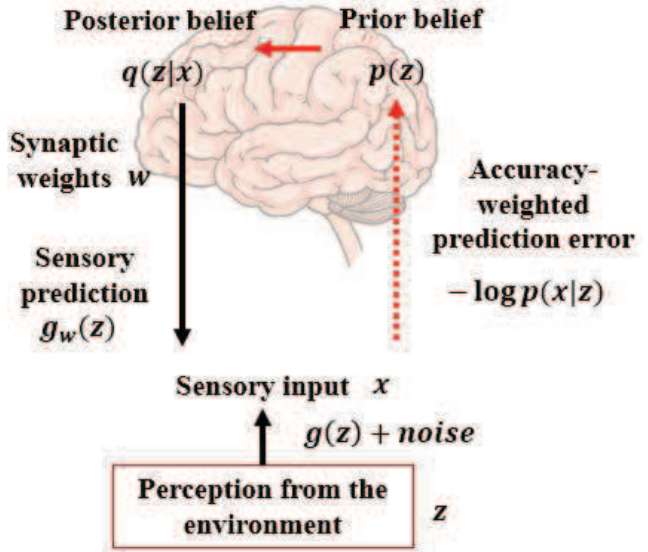}
\caption{
Overview of the free energy principle~\cite{Friston2010nrn}.
}
\label{fig_dpl}
\hspace{10mm}
\end{figure}

\section{Deep Predictive Learning}

This section explains the four strategies that form the basis of the deep predictive learning (DPL) concept, free-energy principle, and predictive coding. Then, as an implementation of the DPL concept, we describe the phases for building a basic predictive model and the three particularly important elements in that phase. In addition, we discussed this concept from the perspectives of embodiment~\cite{Brooks1991intelligence} and dynamic cognitive systems~\cite{Tani1996model}.

\subsection{Main Concept and Policies}

The Deep Neural Network (DNN) framework uses large amounts of data to optimize models, and the goal is for the model to function appropriately, even with new data. However, obtaining an optimal model is usually difficult because it is impossible to determine the state that provides the greatest generalizability. It can be assumed that robots are constantly exposed to unexpected situations because they operate in real environments. Therefore, an approach that assumes that learning is incomplete and considers how to manage prediction errors is required.

There is an information theory of the brain called the free energy principle~\cite{Friston2006jpp,Friston2010nrn} proposed by K. Friston, which can help address the above problem. The free-energy principle states that living things' perception, learning, and behavior are determined by minimizing a cost function called variational free energy (Figure~\ref{fig_dpl}). We summarize this as follows:
\begin{equation}
    FreeEnergy = -\log p(x|z) + D_{KL}[q(z|x)||p(z)] \label{eq_free}
\end{equation}
The above equation corresponds to the negative Evidence Lower Bound in general machine learning models, which is not new. However, by considering the interpretation of the equation and the multiple parameters involved, we can discuss the design of a system with a concrete body. The brain is a predictor with a model $p(z)$ as an a priori belief about the perception $z$ from the environment that causes sensory input $x$. In addition, perception $z$ includes not only the hidden states of exteroceptive sensations, such as vision and touch but also the hidden states of proprioception (e.g., joints) and interoceptive sensations related to kinesthetic and visceral sensations. The brain generates a sensory prediction $g_w(z)$ based on this perception $z$ and obtains a posteriori belief $q(z|x)$ by receiving the current interoceptive sensory input $x$. Subsequently, the brain continuously generates predictions for the next sensory input based on updated posterior beliefs and updates those beliefs. The predicted sensory inputs included the prediction accuracy ( reciprocal of the sensation variance $v_x$). In other words, the first term in equation \ref{eq_free} refers to the prediction error weighted by this accuracy. The second term refers to the influence of the complexity of the prior belief (model) structure.

Strategies for reducing prediction errors are called "predictive coding" and can be classified into four main types. The DPL realizes online motion generation that is tailored to the environment by combining these strategies.

\begin{description}
\item[Training]\mbox{}\\
The basic strategy is to optimize the entire model. The model that generates the sensory predictions is trained, and the internal weight parameter $w$ changes based on the prediction error. However, training a DNN model is expensive and difficult under time constraints. This problem is particularly serious in deep learning approaches because the model size tends to be large. Furthermore, frequent model modifications may be inappropriate depending on the robot's situation.

\item[Modification of Posterior Belief]\mbox{}\\
The second strategy is to modify hypothetical posterior belief $q(z)$. The values of the sensory input were modified in a closed manner based on the predicted perception; as a result, the prediction error of the model was reduced. For example, when humans perceive an environment, they often interpret the real world as an afterthought based on previous brain images. Additionally, the brain is unaware of these errors. This strategy is a method of adapting to real-world changes using predictive biases to create a virtual perception that differs from reality (Section~\ref{sec_generation}).

\item[Active Inference]\mbox{}\\
The third strategy is to use the body to influence the external environment and change the sensory input $x$ to reduce the prediction errors. This operation is called active inference in the context of the free energy principle. These methods include manipulating the environment by oneself or changing viewpoints by changing one's position. This strategy is essential for autonomous agents, such as robots, with bodies in the real world. We implemented behavioral transitions to perform posterior belief and active inference using attractor dynamics embedded in the model's internal state (Sectoin~\ref{sec_imple}).

\item[Modification of Prediction Accuracy]\mbox{}\\
The fourth strategy involves reducing the apparent error by setting the prediction accuracy $1/v_x$ of the sensory input $x$ and multiplying it by the weight of the prediction error. This operation, which increases the precision of sensory input, is associated with directing attention. From the perspective of predictive encoding, this strategy can be interpreted as emphasizing prediction errors and increasing prediction corrections when sensory input is reliable (when attention is directed). However, when sensory input is unreliable (when attention is not directed), the prediction error is ignored, and the inference is maintained. The implementation example corresponds to that presented in Section~\ref{sec_attention}.
\end{description}

\begin{figure*}[t]
\centering
\includegraphics[width=17cm]{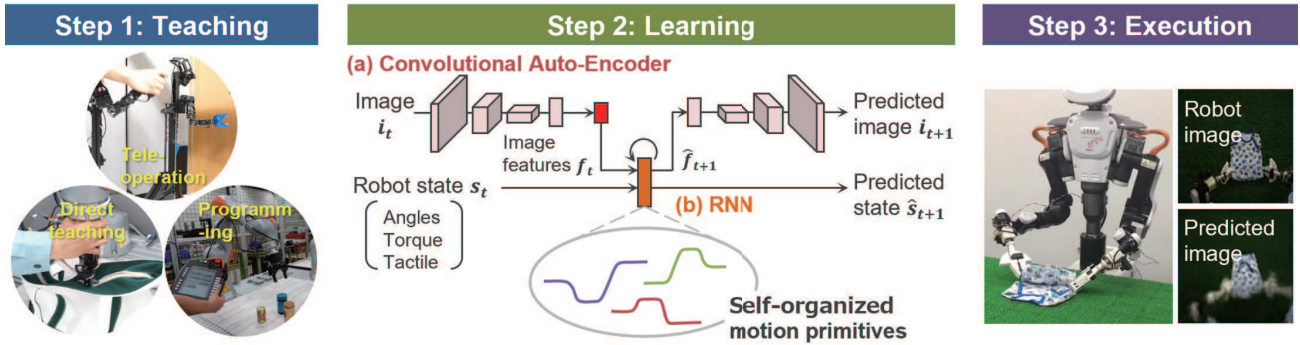}
\caption{
An example of a deep predictive learning framework consisting of a feature extraction unit and a time-series prediction unit.
}
\label{fig_overview}
\end{figure*}

\subsection{Implementation}

\subsubsection{Learning Sensorimotor Dynamics}
\label{sec_imple}

We explain the construction phase of the DPL model to realize the strategies mentioned above, which are divided into training data collection and model learning. Figure~\ref{fig_overview} illustrates the overall flow, including the generation phase described in Section \ref{sec_generation}.

\textbf{Collecting Training Dataset:}
In the DPL concept, a model for generating robot motions is acquired in a data-driven manner by directly using sensorimotor information from the robot (time-series data consisting of motion and sensor information, etc.) as training data. Therefore, the training data must include information on the robot's body and environment interaction. Furthermore, because the quantity and quality of training data greatly affect model performance, it is desirable to collect high-quality demonstration data efficiently.

In the first step, training data are collected by having the robot perform the desired task and record motion information, such as joint angles, and sensor information, such as camera images, at a fixed sampling rate. Typical teaching methods for robot motion include programming, direct teaching, and teleoperation. Describing the motion in advance using a robot programming language is the simplest teaching method; however, the description becomes complicated for long-horizon tasks. However, training data can be obtained without precise modeling or parameter adjustment by having a human remotely control the robot and teaching its motion. In particular, remotely controlling the manipulator from an actual robot's perspective (the Wizard of Oz method) is desirable because it allows the robot to intuitively teach human operational skills for tasks. Workers could interact naturally with the environment by controlling the robot as if it were controlling its own body. The acquired teaching data were expected to include the information necessary for motion learning.

\textbf{Integration Learning of Sensorimotor Dynamics:}
The learning target of our DPL concept is the time-series relationship between sensorimotor information in a system in which the environment and body dynamically interact. The training data did not contain ground truth labels, and the model was trained to predict the next-step robot state using the current robot state as the input. This autoregressive learning eliminates the need for a detailed design of the physical model of the environment, which is required in conventional robotics. In addition, the model can be expressed in terms of dynamics that integrate the environmental recognition and motion generation functions spanning multiple modalities.

We constructed a sensory and motor prediction system by combining multiple DNN models corresponding to the feature extraction and time-series prediction units. The feature extraction unit extracts features from the sensor values acquired by the robot, and the time-series prediction unit learns sensorimotor information that integrates the extracted features and the robot state expressed by joint angles. We published the first model in 2014~\cite{Noda2014multimodal}, and since then, this study has continued to expand. Each part can be connected end to end or learned independently. In this section, we explain the role of each component.

\begin{description}
\item[Feature Extraction Unit:]\mbox{}\\
Sensor inputs that robots obtain from a real environment, such as cameras and tactile sensors, are often high dimensional; therefore, appropriate feature extraction is important. Sensorimotor information is appropriately learned by extracting sensor features that reflect the relationship between the robot body and the manipulated object. The extraction of sensor features affects the prediction performance of the motion information in the time-series learning section, such as when approaching a manipulated object. Therefore, to obtain as detailed information as possible, the feature extractor should handle high-dimensional sensor inputs as is.

Here, as an easy-to-understand example, we consider image data. Raw image data contain much unnecessary information for robot tasks, such as the background, and can become noise for training models. Many models adopted a convolutional autoencoder (CAE) as the feature extractor. In the CAE, low-dimensional features are acquired in the middle layer by performing self-supervised learning to provide an output value equal to the input value. The encoded feature vector provides high-dimensional input information with fewer dimensions, making it robust against environmental noise. Furthermore, the above feature vectors can be narrowed down to more effective information for task learning by introducing an attention mechanism (see Section~\ref{sec_attention}).

\item[Time-series Prediction Unit:]\mbox{}\\
We used a Recurrent Neural Network (RNN) with a recursive connection in the middle layer as the time-series prediction unit. An RNN model reflects temporal dependence and functions as a learning-based controller that inputs the current robot state and outputs the next robot state. The RNN takes the feature vector obtained by the sensor feature extractor and the current robot state (joint angle torque value, gripper value, etc.) as input, outputs the next robot state, and learns to reduce the error between the output and the correct next robot state. A robust model can be constructed by adjusting the ratio of the model output to the previous time as the input. It is also empirically known that predicting sensor values as a subtask contributes to the predictive performance of a model.

Additionally, RNN can retain various dynamics obtained from the time series data learning process as unique internal states~\cite{Tani1998}. When the prediction error increases during robot motion generation, the RNN can switch from learned dynamics to minimize the prediction error (converging to the appropriate dynamics). This property can be used to implement the predictive coding strategies described in the previous section. Furthermore, the composite generation of motions can be realized by representing long-horizon tasks as a combination of dynamics embedded in the RNN. The details are presented in Section~\ref{sec_case_studies}.
\end{description}

\subsubsection{Robot Motion Generation based on Minimizing Prediction Errors}
\label{sec_generation}

As the robot performs a task, it sequentially acquires sensor information. The RNN predicts the robot's state at the next time step by performing forward computation of the model using the input and internal context information at each step. Subsequently, the robot's joints were controlled using the RNN prediction as the target state. The RNN predicts the robot's sensorimotor behavior while sequentially changing the state of each neuron in the context layer by repeating the above process online. One of its advantages is that the calculation time and cost required for motion generation are low because our entire system is composed of light models.

During online motion generation, the manipulated object, robot arm, sensor, and DPL model engage with each other. The model's perceptual predictions (posterior beliefs) were corrected in real-time. As a result of the motion modification, the sensory input was also modified, thereby reducing the model's prediction error. It can be observed that small trajectory corrections and operation switching occur through trial and error~\cite{Ichiwara2022contact,Koma2016ral}. These observations imply that transitions to other behavior attractors occur because of attractions based on prediction errors. What is important here is that trial and error are not random processes but corrections of predicted behavior based on experience. We use these characteristics to accomplish high-dimensional, complex, and long-horizon tasks.

\subsubsection{Hierarchical Structure with Different Time-scales}
\label{sec_timescale}

A hierarchy that considers multiple time scales of a system is also important for predictive learning. In other words, it is a system in which a model that makes predictions on a fast time scale similar to the real world and a model that makes predictions on a much slower time scale exist in parallel. We used a multiple-time-scale RNN (MTRNN~\cite{Yamashita2008emergence}) and LSTM~\cite{Hochreiter1997nc} in the time-series prediction unit and minimized the overall prediction error by the interaction between layers. In the lower layer, which is close to the sensory input, the model predicts fast (short time constant) dynamics, whereas in the upper layer, the model predicts slow (long time constant) dynamics.

In the above model, the sensory prediction errors were absorbed by changes in the state of the lower layer. The upper layer generates predictions based on long-term intentions, giving the lower layer constraints to maintain this state. When the lower layer can no longer absorb a sensory prediction error, it affects the upper layer, causing a state change (change in intention) in the upper layer. This structure, in which the upper and lower layers interact, enables consistent yet adaptive behavior generation~\cite{Suzuki2018ral,Saito2021}.

\subsubsection{Bottom-up and Top-down Attention}
\label{sec_attention}

Next, we describe the attention mechanism involved in modifying prediction accuracy. Generally, attention mechanisms efficiently allocate a model's resources by concentrating the processing area on a specific target. In robotic tasks, it is possible to provide robustness against sudden changes in lighting conditions, placement of multiple objects, and movement of target objects that cause noise. Typically, attention in image processing is generated bottom-up directly from the current input images~\cite{Vaswani2017nips}. For example, methods emphasize brightness, color, direction, and orientation or generate noteworthy keys, queries, and values in self-attention.

In the DPL framework, we are conscious of guiding the attention areas through predictions. By predicting multiple attention points within an image using spatial softmax~\cite{finn2016deep}, we limit the areas that can be attended to in the image and have no choice but to ignore others. We use the attention point and sensorimotor prediction error as a loss function. As a result, the attention point is self-organized, changing depending on one's actions. In addition to bottom-up attention, top-down attention generates attention areas based on the intentions of the upper layer. In \cite{Ito2022integrated}, it was possible to confirm that the background, other than the object being attended to, disappeared in a predicted image generated as a posteriori belief from verbal instruction. Please refer to our related work~\cite{Ito2022integrated,Ichiwara2022contact,Ichiwara2023ral} for details regarding this attention mechanism.

\subsection{Discussion of Proposed Concept}
\subsubsection{Perspectives on Embodiment and Dynamic Cognitive Systems}

In fields based on physicality, such as behavior-based robotics~\cite{Brooks1991intelligence}, intelligent movements are believed to emerge from the coupling of the environment, body, and nervous system. The above fields emphasize the importance of the interaction between the body and the environment. For example, in a passive walking mechanism, the walking motion is generated through the interaction between a well-designed bipedal link model and a slope~\cite{Garcia1996}. Considering the body's flexibility, mechanisms, and sensor placement, it is believed to reserve a variety of actions. At this time, the nervous system only needs to perform minimal functions, and in the example of walking, it is sufficient for the oscillator system to function.

This coupling can provide various states of attraction (attractors) to the nervous system. The DPL concept can be considered a framework in which the above nervous system part is constructed using a deep learning model. For example, there is a discussion of dynamic cognitive systems based on the perspective of dynamical systems proposed by Tani~\cite{Tani1996model}. Here, the brain is not simply considered an input/output system but a predictive system for the world. During interactions with the body, the brain sometimes forms a specific attractor structure with the world; this connection breaks when this discrepancy becomes large. Subsequently, it transitions to a new attractor, sometimes through a phase transition or a chaotic state. As this process is repeated, cognition changes dynamically without converging to a fixed state. Tani et al. realized this process using RNN and a real robot system and have been conducting interesting research for many years, attempting to understand the brain cognitive process~\cite{tani2016exploring}. Although the above study was proposed before the discussion on predictive coding and the free-energy principle, it has many similarities with our DPL concept.

\subsubsection{Contrasting with Reinforcement Learning}

Finally, we discuss the main differences between reinforcement learning and predictive coding. The discussion here is a simple comparison with simple reinforcement learning. As mentioned in the previous section, one of the distinctive characteristics of reinforcement learning is that learning goals can be freely set (for example, scalar values, etc.). The main objective of predictive coding is to minimize prediction errors when the model connects the body and the world. Therefore, setting other learning goals for predictive coding is difficult. Therefore, setting other learning goals for predictive coding will be difficult unless special measures are taken.

In model-based reinforcement learning, a method was proposed for learning a real-world model" by repeating the interactions between actions and states~\cite{Ha2018world}. This method learns actions and visual field images as a sequence of abstracted latent spaces. The learning process is similar to predictive coding in that it targets the world as a prediction target. After learning has progressed sufficiently, the optimal behavior based on the world model can be obtained by searching through trial and error for a behavioral policy that yields the maximum reward from the model.

In contrast, predictive coding assumes that the world is open and that environmental model learning is constantly progressing. When the learning is insufficient, actions expected to reduce future prediction errors (expected free energy) are generated. By repeating the experience of reaching the goal state, the sequence of actions becomes a habit, as a prior belief. Thus, actions with high prediction accuracy were selected. The generated action sequence (virtual expectations) can be considered trial and error in this habituation process. In other words, the behavior learned in this framework is not necessarily optimal for a given objective index. The search that accompanies learning involves optimization and diversification of possible adaptations. The paradox of trying to adapt to the environment but not being able to do so leads to the exploration of various actions.

\section{Case Studies}

This section summarizes the published studies we developed using the DPL concept from two perspectives: scalability by combining multiple motions and application to complex manipulation tasks.

\subsection{Compound Generation of Robot Motions}
\label{sec_case_studies}

As mentioned in the first section, approaches based on large-scale data and trials have high training costs; therefore, it is necessary to perform a wide range of tasks using limited robot motion data. It is an effective approach to express practical tasks by dividing and combining short tasks (subtasks)because many are essentially long-horizon tasks consisting of multiple steps~\cite{Sacerdoti1974ai,Wolpert1998,Schaal2003isrr}. This type of "compound generation of robot motions" has the following advantages:
\begin{itemize}
    \item Using subtasks included in a long-term task for another task (module reuse)
    \item Executing new tasks by combining subtasks (extending functionality)
\end{itemize}
With this background, this section discusses the elements required to combine multiple motions and introduces robot applications using the DPL concept.

\subsubsection{Multi-Module Switching based on Prediction Error}

A simple approach to compound-generating robot motion involves the cooperation of modularized functions~\cite{Haruno2001,Wolpert1998}. Generally, a robotic system consists of multiple modules, and the output is determined by combining them. Even in robot learning, it can be assumed that by combining or switching models with different functions in parallel, the overall task performance can be improved without increasing the computational load. Furthermore, with a distributed system configuration, it is possible to continuously add functionality by adding modules.

One method for achieving this in robot learning is to switch the subtask modules based on the model's prediction error. As mentioned in the previous Section, RNNs can retain the various dynamics obtained from the learning process as unique internal states. When generating robot motion, RNN predicts from the internal state and current sensorimotor information, and the prediction errors tend to increase in untrained states. Therefore, when the prediction error increases during motion generation, the RNN switches from dynamics to minimize the prediction error, which can be used to switch modules. When a large prediction error that cannot be handled in real-time occurs, it cannot be handled by the state transition of the attracter embedded in the individual RNNs. In this case, it is necessary to adjust the overall sensory prediction and motion generation by coordinating multiple modules according to the prediction error.

\begin{figure}[t]
\centering
\includegraphics[width=73mm]{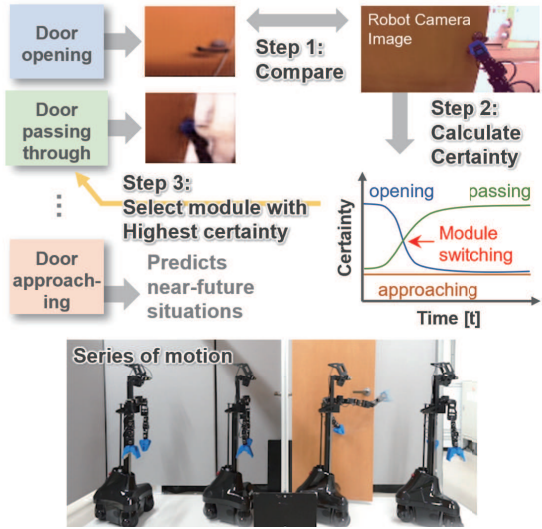}
\caption{
Real-time switching of multiple motion generation models based on prediction errors~\cite{Ito2022sr}.
}
\label{fig_doorrobot}
\hspace{10mm}
\end{figure}

\cite{Ito2022sr} proposed a method that takes actions to reduce prediction errors in door-passing tasks by introducing multiple DPL models and a new mechanism that switches them according to prediction errors (Figure~\ref{fig_doorrobot}). The model was trained for each subtask, such as ``opening a door'' or ``passing through,'' and then combined these models to perform a series of tasks. Each model predicts a near-future camera image from the sensor information. The method calculates the change in the confidence value, which indicates how accurately the robot can work in a real environment over time by comparing it with a real camera image. Furthermore, the robot executes actions that are appropriate for the situation by autonomously selecting the model with the highest confidence. It could handle complex tasks without having to accurately design the timing of the switching motions or the flow of motion because the robot performed these calculations in real time. Based on real robot experiments, we confirmed that generating tasks in response to unlearned door positions and changes in the door panel color and pattern\footnote{https://youtu.be/pdOA-PeYO9Q} is possible.

Contrarily, \cite{Suzuki2021ral} worked on responding to robot states not included in the training data by utilizing multiple control modules with different characteristics. This approach attempted to extrapolate motion datasets (undefined behaviors) that cannot be predicted using a learning-based controller. In the proposed method, we prepared a model-based controller that compensated for the output of the RNN. In addition, the models were appropriately switched based on the error between the predicted and actual trajectories. Assuming that the RNN embeds the most recent information necessary for task execution in its internal state, we train additional tasks to predict past trajectories from the internal state. By adding a new neuron that predicts additional tasks to the middle layer of the RNN, an error recovery functionality can be added without affecting the performance of the main learned task. In real robot experiments, we confirmed that, by combining this method with the dynamic design of the RNN internal state described in the next section, it is possible to recover errors in response to multiple types of disturbances (environmental contact and changes) from collaborative workers.

In general robot systems, it is necessary to separately program responses to unexpected situations, which require significant development costs and adjustment work. In addition, when an unexpected disturbance occurs in the work environment, the robot must take time to recognize the situation and replan, which causes it to move slowly. Our proposed methods~\cite{Ito2022sr,Suzuki2021ral} can respond to sudden situational changes and changes in operating procedures by switching multiple controllers according to prediction errors and are expected to solve the above problems.

\subsubsection{Switching of Multiple Behaviours by Designing the Dynamical Systems}

Next, we discuss the dynamic system of the RNN as a method for handling local composite generation within a single module. The method described in the previous section focuses on switching between modules; however, considering scalability, it is desirable to have a mechanism that can switch operations within individual modules. This need arises because when a task is subdivided, each module must handle multiple subtasks~\cite{Nair2019iclr,Yu2019iros}. Previous studies~\cite{Yin2021iclr,Garrett2021cras} also investigated motion planning that combines subtasks. Expressing their dynamics in a combined format within a module is necessary to connect multiple robot motions. Long-term task operations can be executed in response to environmental changes by selecting the appropriate dynamics of the RNN according to the sensory input. We implemented the above by explicitly embedding the attractor dynamics in the subtask units into an RNN.

\begin{figure}[t]
\centering
\includegraphics[width=\columnwidth]{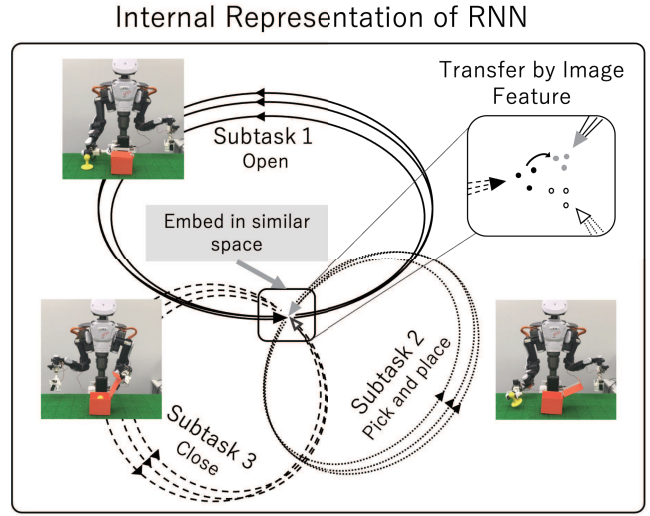}
\caption{
Embedding attractor dynamics into the internal state of RNN~\cite{Kase2018icra}.
}
\label{fig_kase_attractor}
\hspace{10mm}
\end{figure}

In \cite{Kase2018icra}, to combine multiple subtasks, we intentionally embedded an attractor structure inside the model using a loss function in the internal state of the RNN (Figure~\ref{fig_kase_attractor}). The attractor switches its dynamics to reduce the prediction error by adding a constraint loss to obtain a shared state that can be switched between the attractors of each subtask. In this experiment, combining multiple subtasks for a put-in-box task generated motions in an arbitrary order of subtasks\footnote{https://youtu.be/QwogmDqOxfE}. Furthermore, we expanded the loss function of the internal state of RNN and designed an arbitrary attractor structure to increase the degrees of freedom in switching postures~\cite{Kase2019iros}.

\setlength\textfloatsep{5pt}

\cite{Suzuki2018ral} introduced a hierarchical structure of task instruction input and RNN to learn how to switch subtasks during a long-horizon task. Inputting a task instruction vector to the designed point attractor enables the response to motion switching in ambiguous situations that cannot be determined based on sensor information alone. In the experiment, we verified a clothes-folding task that consisted of multiple subtasks and confirmed that the robot could perform it according to the instructions. Furthermore, in the internal state of the RNN, instructions were embedded in the long-term timescale layer, and motion was embedded in the short-term timescale layer as an attractor structure. These embeddings indicated that the task context, language, and image inputs were appropriately combined.

Although these motion combination/switching methods using attractor structures require the design of switching postures for subtasks, they can incorporate human intentions to some extent into the RNN output. The robot is expected to be able to switch its motions more flexibly, combined with grounding, using the language described below.

\subsubsection{Language-Conditioned Motion Generation}

Grounding each segmented subtask and language expression is necessary to express more primitive action combinations. The RNN expresses internal dynamics with the semantic and syntactic structure of the language system as a constraint by learning that the given language instructions and individual tasks are semantically grounded. This internal representation can be extended to cases where actions are sequentially generated while communicating with collaborators. Based on the above idea, we aim to inductively express the relationship between motions and symbols contained in sensorimotor information~\cite{Harned1990pd} within a model.

\textbf{Grounding of Language and Robot Motion:}
In general, the correspondence between language, which is a discrete expression, and robot motion in a continuous space changes depending on the past context and situation on the spot. For example, even with simple instructions such as "Hit the bluebell," the action to accomplish this will vary depending on where the bluebell is placed, what position the robot is in, etc. Robots must acquire internal representations that can be generalized, even in unknown situations, to achieve such ambiguous and one-time grounding each time.

\begin{figure}[t]
\centering
\includegraphics[width=\columnwidth]{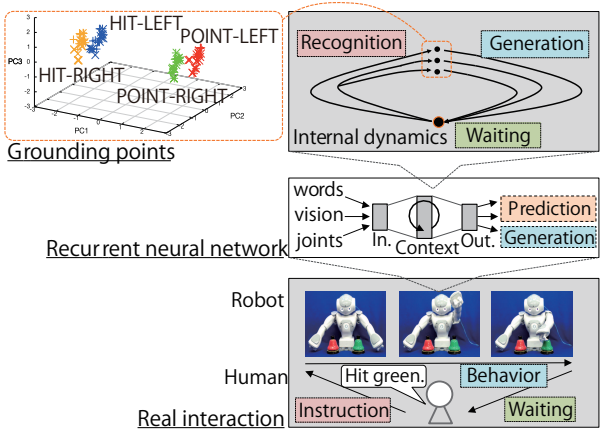}
\caption{
Grounding of language instructions and motion generation in the designed internal dynamics of RNN~\cite{Yamada2016fn}.
}
\label{fig_language_reduce}
\hspace{10mm}
\end{figure}

Our DPL framework enables embedding an understanding of language instruction and motion into a set of attractor structures. \cite{Yamada2016fn,Yamada2017fn} acquire cyclic attractors that express the above episodic repetition as the internal dynamics by having RNN predict and learn a sequence that repeats multiple episodes of waiting, verbal instructions, and robot motion generation. (see Figure~\ref{fig_language_reduce}). When a verbal instruction and a camera image are input to the RNN at a fixed-point attractor, the internal state of the RNN changes in a branching manner according to the local situation. After the language input was completed, robot motions were generated according to the meaning of the instructions in the environment. Once the motion was complete, the internal state converged to a fixed point and waited for the next instruction. We demonstrated the generalization ability to generate appropriate actions for new situations without having to teach every combination of environmental conditions and verbal instructions.

\setlength\textfloatsep{5pt}

Furthermore, extending this model to a bidirectional conversion model that includes conversion from task instructions to actions and from actions to linguistic explanations is possible. In \cite{Yamada2018ral}, we prepared two autoencoders, one for language and one for sensorimotor sequences. We performed learning by setting constraints such that the feature vectors of the corresponding sentences and motions were close to each other. After the training phase, the model achieved a two-way translation from instructions to robot motions and robot motions to linguistic explanations via feature vectors. \cite{Toyoda2021ral,Toyoda2022ral} added that model a word distributed representation conversion module that utilizes the Word2vec model, allowing it to perform appropriate operations even for instructions that include new words similar to trained words.

\begin{figure}[t]
\centering
\includegraphics[width=\columnwidth]{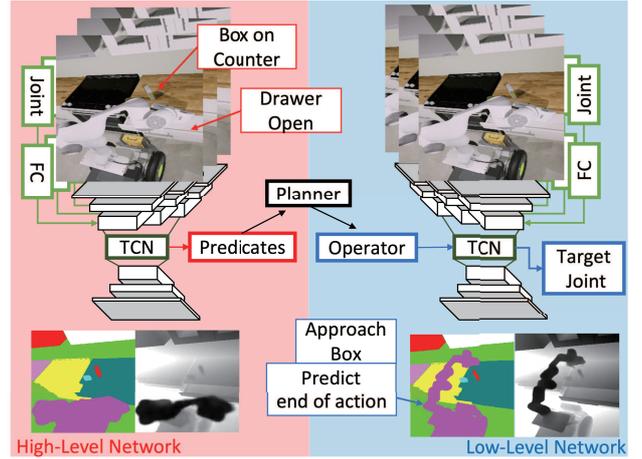}
\caption{
Grounding with a generic planning language and extracted image features~\cite{Kase2020icra}.
}
\label{fig_kase2022icra}
\hspace{10mm}
\end{figure}

\textbf{Integration with Task Planning:}
In the above studies, we introduced methods to provide instructions in the language form and integrate the robot's motions. Linguistic instructions are given word by word; however, general-purpose planning description languages (e.g., planning domain description language: PDDL) used in conventional automatic planning methods can be used as task instructions. \cite{Kase2020icra} proposed a learning method to connect descriptive language that defines the environment and images (Figure~\ref{fig_kase2022icra}). They designed motions using an automatic planning method based on discrete symbolic representations extracted from images. Although conventional PDDL makes it difficult to respond to planned environmental changes, our method makes it possible to determine the robot's actions dynamically.

In recent years, large language models (LLM~\cite{Zhao2023}) have been actively applied in robot motion generation~\cite{Shridhar2021a,Liang2022a,Brohan2022a,Driess2023a}. Especially in task planning, it is possible to convert linguistic instructions into multiple subtask instructions in table format~\cite{Ahn2022,huang2022language} and code format~\cite{Vemprala2023,codeaspolicies2022}. Although the studies introduced in this section deal with limited language formats, we plan to expand them by incorporating LLM technology~\cite{Hori2024}.

\subsection{Application to Various Robot Tasks}
\subsubsection{Deformable Object Manipulation}

We verified the versatility of the DPL concept by applying it to difficult tasks. A typical example is the manipulation of deformable objects. Manipulating flexible objects such as cloths and strings that change shape upon contact makes it difficult to describe the entire situation in a program. \cite{Koma2016ral} generalized the motion of folding a towel by performing integrated learning based on image-motion information acquired through the robot's work experience. As a result, the robot could repeatedly perform tasks corresponding to object position shifts and different objects (such as books) using limited training data. Unlike conventional methods, our method does not require high-cost processing such as modeling or image processing, and it is possible to generate motion at a natural speed\footnote{https://youtu.be/YH1TrL1q6Po}. Our concept does not depend on tasks or robot hardware; therefore, by implementing each function as a component, we can easily reuse the implemented system in response to changes in requirements~\cite{Kanamura2021sensors}. We realize the in-air knotting task~\cite{Suzuki2021iros} and the buttoning task~\cite{Fujii2022sii} using the same model in previous studies.

\begin{figure}[t]
\centering
\includegraphics[width=\columnwidth]{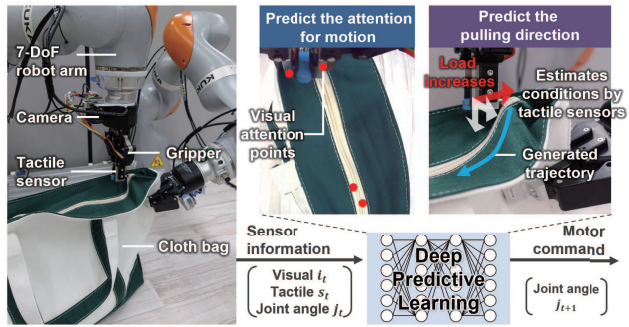}
\caption{
Generated bag opening task using attention mechanism with vision and tactile sensors~\cite{Ichiwara2022contact}.
}
\label{fig_attention_bag}
\hspace{10mm}
\end{figure}

It has been confirmed that including detailed object information, such as proximity and tactile sensors, as input to the model improves the prediction performance. \cite{Ichiwara2022contact} introduced an attention mechanism that extracted information (such as the position and direction of the object) from the robot's visual and tactile information (Figure~\ref{fig_attention_bag}). In this task, generating an appropriate trajectory in real-time is necessary according to the state of the bag and robot hand. The game starts by grabbing randomly placed zippers so that the gripping position and posture constantly change. The number of training sequences was 36, which is extremely small compared with general learning methods. The experiment confirmed that the model's perceptual predictions (predicted images, tactile states, and motions) were corrected in real time using an attention mechanism.

Touch also plays an important role in the emergence of diversity generated from exploration based on intrinsic motivation. Predictability and operability are the criteria for learning; however, they do not indicate the possibility of further exploration. We also proposed searching for new actions based on tactile perceptual information~\cite{Mori2020} and active inference based on the internal characteristics of the manipulated object~\cite{Saito2021,Saito2023a}. These studies address the next important issue to be solved in our DPL concept from the perspective of the emergence of various movements based on exploration.

\setlength\textfloatsep{5pt}

\subsubsection{Neuro-Robotics Experiments}

We also used a similar hierarchical predictive model for robot simulation of neurodevelopmental disorders. We reproduced the functional disconnection between layers~\cite{idei2021paradoxical} and the imbalance between excitability and inhibition in neurons~\cite{idei2020homogeneous}. These studies show that when a real robot causes an abnormality in sensory accuracy or even upper-layer prediction accuracy, it is observed as a behavioral abnormality. Although this paper does not go into detail, the studies mentioned above are significant because they extend the DPL concept from a perspective not limited to task learning.

\section{Conclusion}

In this study, we introduce deep predictive learning, a concept of robot motion learning. We also summarize the background technology of predictive coding, related studies, and robotic applications of the proposed concept. The three robot application examples described regarding the composition of multiple motions are as follows: switching between multiple modules based on prediction errors, designing dynamic systems inside the DNN model, and acquiring an integrated language-motion representation. In addition, we demonstrate their applicability to various robotic tasks, including flexible object manipulation. Although each method has been developed separately, many common technical elements are expected to be integrated in the future. However, there are challenges in improving the accuracy and applying the system to more difficult tasks. We plan to revisit the proposed concept from the perspective of cognitive robotics the concept's origin, and apply the method to long-term and more diverse tasks in line with real-world problems.

\section*{ACKNOWLEDGMENT}
The authors would like to thank Dr. Hayato Idei for his comments on the description of predictive coding in this manuscript.
This work was supported by JST Moonshot R\&D Grant Number JPMJMS2031, JST ACT-X GrantNumber JPMJAX190I, Hitachi Ltd. Japan.

\bibliographystyle{ieeetr}
\bibliography{refs2}

\end{document}